\documentclass[10pt,twocolumn,letterpaper]{article}

\usepackage[pagenumbers]{cvpr} 

\usepackage{graphicx}
\usepackage{amsmath}
\usepackage{amssymb}
\usepackage{booktabs}
\usepackage{multirow}
\usepackage[font={normal}]{caption}
\usepackage[pagebackref,breaklinks,colorlinks]{hyperref}
\usepackage[accsupp]{axessibility}

\usepackage[capitalize]{cleveref}
\crefname{section}{Sec.}{Secs.}
\Crefname{section}{Section}{Sections}
\Crefname{table}{Table}{Tables}
\crefname{table}{Tab.}{Tabs.}

\begin{document}
\title{NeMF: Inverse Volume Rendering with Neural Microflake Field}

\author{
Youjia Zhang$^{1}$
\quad
Teng Xu$^1$
\quad
Junqing Yu$^1$
\quad
Yuteng Ye$^1$
\quad
\\ Junle Wang$^2$
\quad
Yanqing Jing$^2$
\quad
Jingyi Yu$^3$
\quad
Wei Yang$^{1*}$
\quad
\\[1.5mm]
$^1$Huazhong University of Science and Technology
\quad
$^2$Tencent
\quad
$^3$ShanghaiTech University
}

\vspace{-16pt}
\twocolumn[{%
\renewcommand\twocolumn[1][]{#1}%
\maketitle
\begin{center}
    \centering
    \captionsetup{type=figure}
    \includegraphics[width=1.0\textwidth]{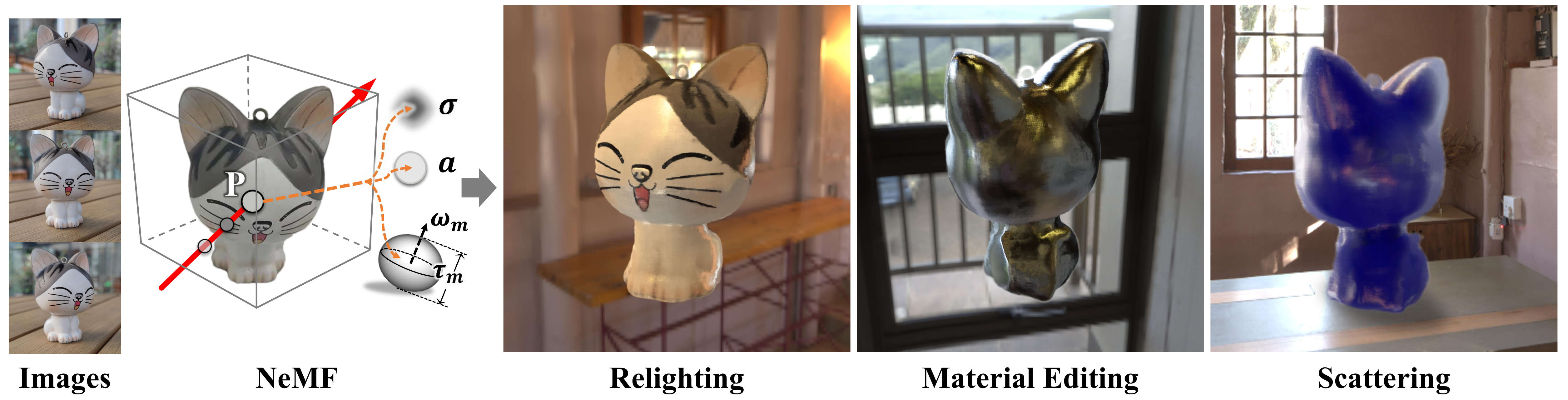}
    \captionof{figure}{We present the Neural Microflake Field (NeMF) for inverse volumetric rendering from multi-view images under unknown natural illumination. NeMF represents the scene as a microflake volume, in where light reflects or scatters at each spatial location according to volume density, microflake roughness and normal. The optimized NeMF enables high-quality relighting, material editing, synthesize volume scattering effects (shadow is imposed using the preset shadow effect in Microsoft PowerPoint for better exhibition) and etc.}
 \label{fig:teaser}
\end{center}%
}]

\begin{abstract}
Recovering the physical attributes of an object's appearance from its images captured under an unknown illumination is challenging yet essential for photo-realistic rendering. Recent approaches adopt the emerging implicit scene representations and have shown impressive results.However, they unanimously adopt a surface-based representation,and hence can not well handle scenes with very complex geometry, translucent object and etc.In this paper, we propose to conduct inverse volume rendering, in contrast to surface-based, by representing a scene using microflake volume, which assumes the space is filled with infinite small flakes and light reflects or scatters at each spatial location according to microflake distributions. We further adopt the coordinate networks to implicitly encode the microflake volume, and develop a differentiable microflake volume renderer to train the network in an end-to-end way in principle.Our NeMF enables effective recovery of appearance attributes for highly complex geometry and scattering object, enables high-quality relighting, material editing, and especially simulates volume rendering effects, such as scattering, which is infeasible for surface-based approaches. \\
\end{abstract}
\vspace{-1em}
\let\thefootnote\relax\footnotetext{$^*$Corresponding author: Wei Yang.}

\vspace{-30pt}
\section{Introduction}
\label{sec:intro}
Inverse rendering refers to the process of recovering an object's physical attributes related to its appearances, including shape, reflectance and illumination, from its image observations. The above physical attributes play a vital role in graphic applications that require physically reasonable realism. 
The problem is highly ill-posed due to the complication of object geometries, material and varieties of illuminations. 
It becomes even more difficult when the images are captured under an unknown illumination condition. The typical practice is to represent the object as surfaces and then solve for the Spatially-Varying Bidirectional Reflectance Distribution Functions (SVBRDF) at each ray-surface interaction.
Traditional approaches rely on restrictive assumptions~\cite{deepsvbrdf, deepbrdf, efficientbrdf, image-based, planned} or sophisticated capture systems, such as light-stages~\cite{debevec2000acquiring}, co-located flashlight and camera setup~\cite{IRON}, and etc.



More recent works explore the implicit scene representations, e.g., radiance field and signed distance functions~\cite{yariv2020multiview, jiang2020sdfdiff, park2019deepsdf, maier2017intrinsic3d, zollhofer2015shading}, and achieve promising results. They exploit the geometry, reflectance, or visibility recovered by implicit models as initial estimates for solving the ill-posed inverse rendering problem. Notably, NeRD~\cite{nerd} adopts a two-stage estimation strategy by first predicting the sampling pattern and albedo, and then performing per-ray SVBRDF decomposition. NeRFactor~\cite{nerfactor} applies hard surface approximation on NeRF geometry and recovers neural fields of surface normals, light visibility, albedo and SVBRDFs. Zhang et al.~\cite{indirect} represent the scene geometry as a zero-level set and recover spatially-varying indirect illumination for more accurate inverse rendering.  Nevertheless, they either rely on a surface-based representation or need to extract geometry from a volume representation first for the subsequential reflectance estimation. This usually leads to a multi-stage refinement framework, and the performance depends heavily on qualities of the recovered geometries.

\begin{figure}[t]
	\centering
	\includegraphics[width=1.0 \linewidth]{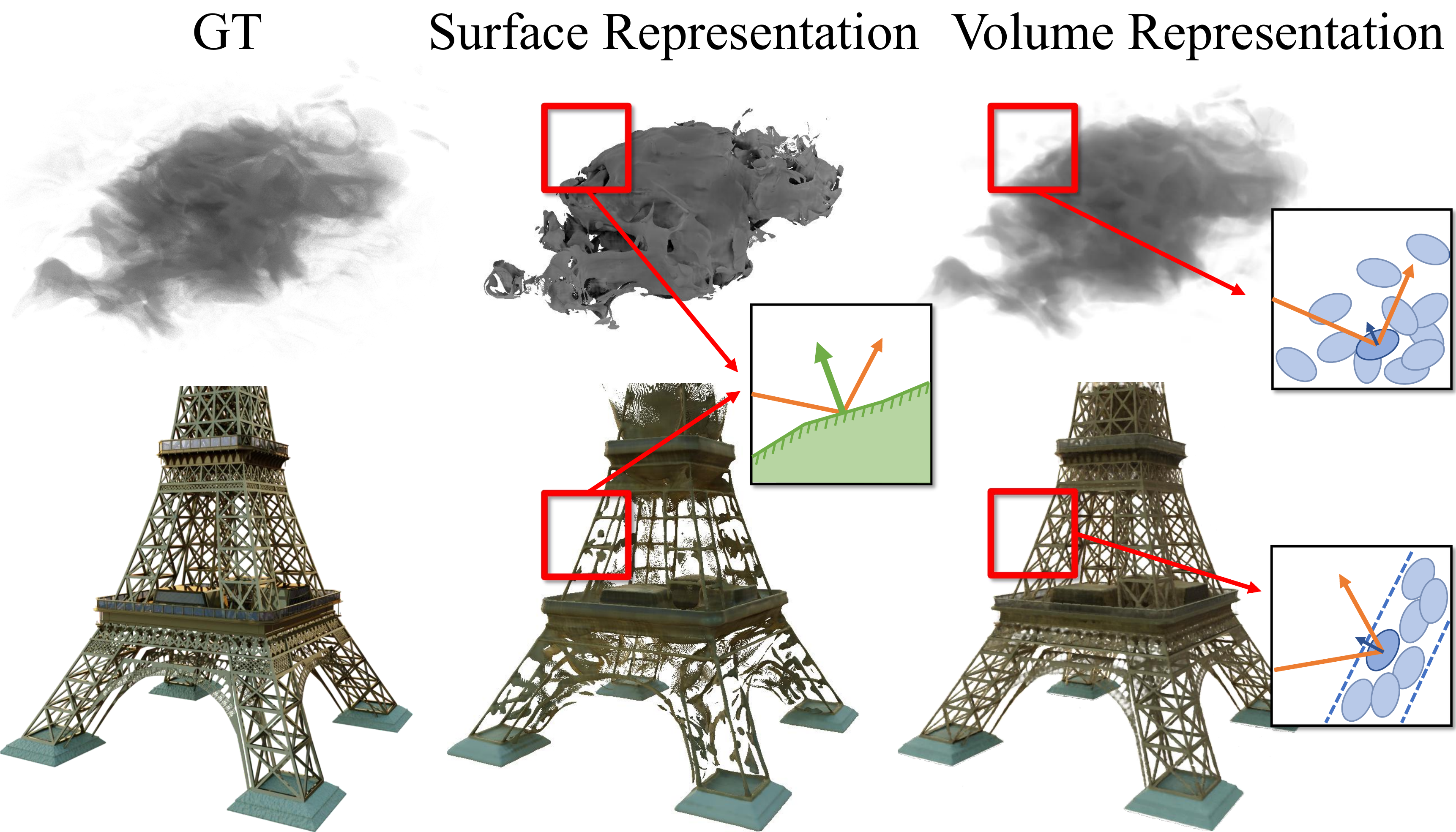}
	
	\caption{
        A surface-based representation can not handle scattering materials (e.g., cloud) and very complex geometry (e.g., Eiffel Tower). In contrast, the microflake volume can both handle a surface-like behavior by applying higher density inside the object and low density outside, and a volumetric object.
    }
    \label{microflake}
\end{figure}

Recall that the seminal work of NeRF~\cite{nerf} adopts a radiance volume representation and enables photorealistic rendering without explicit geometry modeling. In essence, NeRF assumes the space is filled with infinite small particles that emit radiation. In this paper, we faithfully extend the volumetric setup by replacing the particles with oriented flakes, which reflect or scatter light according to space occupancies and materials~\cite{sggx}.The interaction of light with a collection of microflakes in the volume is described by the microflake phase function, which is determined by the ellipsoidal distribution of normals (NDF) and further parameterized by the microflake normal direction $\omega_m$ of and roughness $\mathbf{\tau}_m$, as shown in Fig.~\ref{fig:teaser}. Such representation can also simulate surface-like object with denser flakes inside the object, while sparser outside, as shown in Fig.~\ref{microflake}.With this Neural Microflake Field (NeMF) representation, we propose to conduct inverse volume rendering, to tackle the overly dependency on geometry problem of existing methods.We start from the vanilla NeRF model, add one additional MLP branch for estimating the mircoflake normal $\omega_m$. As for the microflake roughness $\tau_m$, we use a U-shaped MLP network to first encode the material to a feature vector, apply sparsity constraint and then decode it back into albedo $a$ and roughness $\tau_m$. To estimate the density, albedo, and microflake distributions, we develop a differentiable microflake volume renderer and use the photometric loss between the rendered and groundtruth images for supervision. We evaluate the proposed method on both synthetic and real datasets. The experimental results show that our approach outperforms existing methods in terms of rendering quality, and is able to recover scenes with complex geometry and translucent objects. Moreover, our NeMF not only enables effective relighting and material editing but also allows for simulating volume scattering, as shown in Fig.~\ref{fig:teaser}.

\begin{figure*}[t]
	\centering
	\includegraphics[width=1.0 \linewidth]{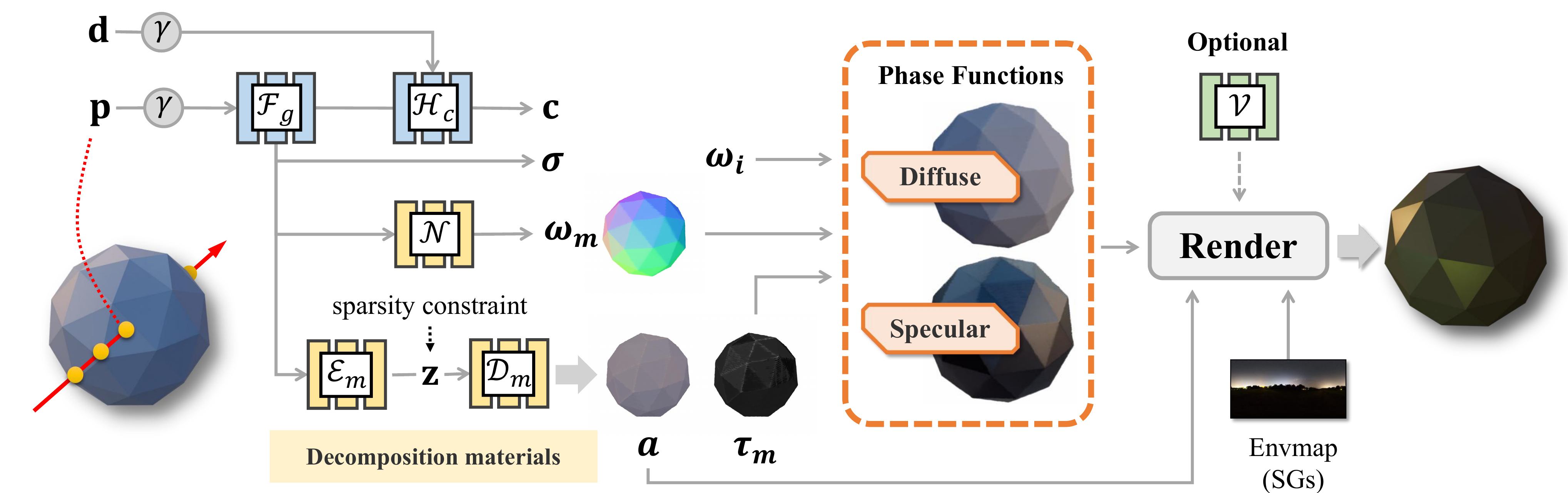}
	
	\caption{
		The framework of our NeMF, where we use multiple branches of MLP networks to predict the volume density $\sigma$, albedo $a$, microflake normal $\omega_m$ and roughness $\tau_m$ respectively. The predicted volume parameters are then sent to a differentiable microflake volume render for rendering.
	}
	\label{forward rendering}
\end{figure*}

\section{Related Work}
\label{sec:rw}

Our work is closely related to research in inverse rendering and implicit scene representations.

\noindent \textbf{Inverse rendering.} Inverse rendering is a vital problem in both computer vision and computer graphics. One popular way to resolve inverse rendering problems is to use strong scene priors~\cite{yu2019inverserendernet, wei2020object, sengupta2019neural, sang2020single, lichy2021shape, li2018learning, li2020inverse, barron2014shape, nerd, nerfactor}. This type of approach recovers intrinsic image properties to infer objects under novel views or illuminations. Another class of inverse rendering methods heavily depends on additional observations.
Some of them require input images with known camera parameters~\cite{yariv2020multiview, jiang2020sdfdiff, park2019deepsdf, maier2017intrinsic3d, zollhofer2015shading} or under known lighting source~\cite{schmitt2020joint, bi2020neural, nam2018practical, bi2020deep, bi2020deepref}, while others take 3D geometry obtained from active scanning~\cite{zhang2021neural, park2020seeing, schmitt2020joint, lensch2003image, guo2019relightables}, proxy models~\cite{chen2020neural, gao2020deferred, georgoulis2015gaussian, dong2014appearance}, silhouette masks~\cite{xia2016recovering, godard2015multi, oxholm2014multiview}, or multi-view~\cite{goel2020shape, philip2019multi, nam2018practical} stereo as a precondition.

Recent work extends inverse rendering to more flexible scene conditions through implicit neural representation and achieves promising results. As exemplary, NeRFactor~\cite{nerfactor} and PhySG~\cite{physg} decompose scenes into geometry (NeRFactor extract geometry from NeRF and PhySG relies on SDFs), material and lighting under unknown illumination. PhySG only handles static illumination. NeRFactor models spatially-varying reflectance with low-frequency BRDFs. NeRV~\cite{nerv} considers indirect illumination with known direct illumination. InvRender~\cite{indirect} recovers indirect illumination based on geometry recovered by SDF methods. Most existing approaches either use surface-based representations, e.g., SDF, or extract geometry from volume representations, such as NeRF. Our approach fully relies on volume representation, and hence do not need to recover geometry explicitly. The most related approach to ours is NeRD~\cite{nerd}, which calculates the density and SVBRDF parameters for each scene point. However, it accumulates the SVBRDF parameters along a ray as the final material parameter of the ray/pixel, resembling a ray-based reflectance decomposition, which is fundamentally different from our approach.
Different from the prevalence of surface-based inverse rendering problems, research on inverse volume rendering is limited. Existing inverse volume rendering approaches either focus on recovering translucent materials using special acquisition device~\cite{gkioulekas2013inverse}. or efficiently differentiating the radiative transfer equation~\cite{vicini2021path, zhang2021path, nimier2022unbiased}.

\noindent \textbf{Implicit neural representation.} 
Implicit neural representation is a recent trend that encodes the scene implicitly via a neural network. Typically, they adopt a coordinate network to infer properties related to scene geometry per scene point. NeRF~\cite{nerf} achieves particularly satisfactory performance, enabling photo-realistic novel view synthesis by using MLPs to represent scenes as radiance fields. While NeRF achieves scene representation based on volume, DeepSDF~\cite{park2019deepsdf, sitzmann2020metasdf} proposes to use a coordinate networks to ecode the zero levelset surface. Occupancy Networks~\cite{mescheder2019occupancy} predict the occupancy of each scene point via a coordinate network, the geometry surface then is the place where occupied and vacant exchange. Though these representations both work well for novel view synthesis, they do not model the light transport or reflection in the volume, and hence infeasible for relighting tasks.

\noindent \textbf{Volume Rendering} 
Volume rendering accumulates radiance along the ray passing the volume~\cite{drebin1988volume, niemeyer2020differentiable}. Analogous to the microfacet in modeling surfaces, Jakob et al.~\cite{jakob2010radiative} propose the `microflake theory' to simulate volumetric scattering with arbitrary microstructures. 
Extended from the microflake framework, the Symmetric GGX (SGGX)~\cite{sggx} provides the ability to represent microstructure with lightweight storage as well as to represent specular and diffuse microflakes in a unified manner. 

\section{Neural Microflake Field}

In this section, we describe our Neural Microflake Field (NeMF) which introduces the microflake volume model to implicit scene representations. 

\subsection{The Microflake Distribution}

The microflake distribution is designed for modeling spatially-varying properties using oriented non-spherical flakes for simulating light transportation inside a volume. We follow the theory proposed by Heitz~\cite{sggx}. The interaction of light with a collection of microflakes is described by a phase function determined by their distribution of normals (NDF). The NDF serves as a weighting function to scale the radiation transportation in directions. More specifically, the theory simulates the NDF of a collection of microflakes using an ellipsoid, with the ellipsoid's normal $w_m$ and projected areas $\tau_m$ onto normal's orthogonal tangent directions $\mathbf{\omega}_x$ and $\mathbf{\omega}_y$. Hence the ellipsoid can be parameterized by $w_m$ and $\tau_m$ according to a 3 × 3 symmetric positive definite matrix $S$:
\begin{equation}
S=(\omega_x,\omega_y,\omega_m) 
\begin{pmatrix}
\tau_m^2 & 0 & 0 \\
0 & \tau_m^2 & 0 \\
0 & 0 & 1
\end{pmatrix}
(\omega_x,\omega_y,\omega_m)^T
\end{equation}

\noindent We can model the probability of normals in any given direction $\omega$ as on the ellipsoid surface:

\begin{equation}
D(\omega) = \frac{1}{\pi \sqrt{| S |} (\omega^T S^{-1} \omega)^2}
\end{equation}


With the microflake distributions defined by $D(\cdot)$, we can model the appearance of diffuse materials and specular materials through phase functions, which measure the attenuation given an input light direction $\omega_l$ and viewing direction $\omega_i$. 

\noindent \textbf{Diffuse Phase Function} The phase function of diffuse microflakes then is the integral of attenuation according to angles between normals and incoming and outgoing light directions:

\begin{equation}
f_p^d(\omega_i, \omega_l) = \frac{1}{\pi \tau (\omega_i)} \int_{\Omega} \langle \omega_l, \omega \rangle \langle \omega_i, \omega \rangle D(w) d_{w}
\label{diffpf}
\end{equation}
\noindent where $\langle \cdot \rangle$ denotes the dot product, $\Omega$ is the hemishpere centered at $\omega_m$, and $\tau (\omega)$ is the projected size of the ellipsoid along direction $\omega$.

\noindent \textbf{Specular Phase Function} The phase function for a specular microflakes then is only related to the half angle $\omega_h$ as:

\begin{equation}
f_p^s(\omega_i, \omega_l) = \frac{D(\omega_h)}{4 \tau(\omega_i)}
\label{specpf}
\end{equation}

Notice Eqn.~\ref{diffpf} and Eqn.~\ref{specpf} define the reflectance at a certain position in space. The volumetric rendering process requires the integration of light reflected by microflakes along the target ray, and we will explain the process in Sec.~\ref{sec:rwn}.

\subsection{Implicit Microflake Field}
\label{sec:imf}

We can consider the radiance field representation in NeRF as a volume of particles with view-dependent radiance. Such representation enables photorealistic view synthesis while prohibiting inverse rendering as it's incapable to model light transportation. We find the microflake model can serve as the natural extension of the radiance field for radiation transportation modeling. Recall in the previous section, we can use the microflake normal $\omega_m$ and projected area as roughness $\mathbf{\tau}_m$ to fully represent the microflake distributions. And we adopt a coordinate-based neural network $\Phi$ to estimate parameters related to microflake models for each scene point $\mathbf{p}$, as:

\begin{equation}
\Phi: \mathbf{p} \rightarrow (\sigma, a, \mathbf{\omega}_m, \tau_m)
\end{equation}

\noindent where $\sigma$ is the volume density and $a$ is the albedo.

However, using a single network to regress all parameters is impractical. We observe that $\sigma$ and $\omega_m$ are related to scene geometry, and the NeRF structure can provide good estimates. While the albedo $a$ and $\tau_m$ correspond to object appearance, we can enforce sparsity constraints according to~\cite{indirect}. Hence we start with the NeRF network $\mathcal{F}_g$ and $\mathcal{H}_c$, add an additional branch $\mathcal{N}$ for estimating the microflake normal, and we use a U-shaped MLP structure with encoder $\mathcal{E}_m$ and decoder $\mathcal{D}_m$ for mapping appearance into sparse latents $\mathbf{z}$. We have:

\begin{equation}
\Phi = \{ \mathcal{F}_g, \mathcal{H}_c, \mathcal{N}, \mathcal{E}_m, \mathcal{D}_m \}
\end{equation}
\noindent where:
\begin{equation}
\begin{gathered}   
\mathcal{F}_g: \mathbf{p} \rightarrow \sigma ; \,\, 
\mathcal{F}_g + \mathcal{H}_c: (\mathbf{p}, \mathbf{d}) \rightarrow \mathbf{c}; \,\, \mathcal{F}_g + \mathcal{N}: \mathbf{p} \rightarrow \mathbf{\omega}_m \\
\mathcal{F}_g + \mathcal{E}_m: \mathbf{p} \rightarrow \mathbf{z} ; \,\, \mathcal{D}_m: \mathbf{z} \rightarrow (a, \tau_m)
\end{gathered}
\end{equation}

\noindent $\mathbf{d}$ is the query ray direction. $\mathbf{c}$ is the color. The implicate networks used for modeling the microflake field is shown in Fig.~\ref{forward rendering}. 

\subsection{Rendering with NeMF}
\label{sec:rwn}
Our NeMF represents the scene as a distribution of microflakes at every point in space. Rendering with the microflake volume follows the principles provide in~\cite{sggx}, i.e, for a given ray $\mathbf{r}$ with direction $\mathbf{d}$ passing through the volume, the resulting pixel intensity is integral of radiance along $\mathbf{r}$.

\begin{equation}
C(\mathbf{r})=\int_{t_n}^{t_f} \eta(t) \sigma(\mathbf{r}_t) \mathbf{\nu}(\mathbf{r}_t, \omega_i)  d_t
\label{eq:volrender}
\end{equation}

\noindent where $\mathbf{r}_t$ means the point on $\mathbf{r}$ at $t$, $\eta(t) = \exp (-\int_{t_n}^t \sigma(\mathbf{r}_s) d_s)$ is the weight,  $\nu(\mathbf{r}_t, \omega_i)$ is the radiance at point $\mathbf{r}_t$ in the direction $\omega_i$ of $\mathbf{r}$, where $\omega_i = -\mathbf{d}$ is a unit vector pointing from a point in space to the camera. For NeRF, the radiance is the view-dependent color to be predicted by the network. In our NeMF, the radiance at a scene point $x$ refers to its transported radiation, and is an integral of all incoming light that reaches at $x$, attenuated by the phase function given outgoing light direction (i.e., the viewing direction $\mathbf{d}$). Hence, we can calculate $\nu(\mathbf{r}_t, \omega_i)$ with a known phase function $f_p$ as:
\begin{equation}
\nu(\mathbf{r}_t, \omega_i, f_p)= \alpha \int_{\Omega} f_p[\omega_i, \omega_m(\mathbf{r}_t), \omega_l] \cdot L(\mathbf{r}_t, \omega_l) d_{\omega_l}
\label{eq:radatx}
 \end{equation}

\noindent where $L(\mathbf{r}_t, \omega_l)$ is the light intensity reaches $\mathbf{r}_t$ in direction $\omega_l$. Recall that we have separated the microflake phase functions into diffuse and specular components, hence we have the final formula for $\nu$ as:

\begin{equation}
\nu^*(\mathbf{r}_t, \omega_i) = \nu(\mathbf{r}_t, \omega_i, f_p^d) + \nu(\mathbf{r}_t, \omega_i, f_p^s)
\label{eq:radcomb}
\end{equation}

Combining Eqn.~\ref{diffpf}, \ref{specpf}, \ref{eq:radcomb}, \ref{eq:radatx}, and~\ref{eq:volrender}, we obtain the rendering equations of our NeMF. However, in practice we have to adapt the discrete data from sampling, we replace integral with summation over discrete sampling positions. Then another issue to be mindful is the phase functions measure light attenuates according to incoming and outgoing directions. When we calculate $\nu^*(\cdot)$, we need to conduct  importance ray sampling of light directions, which is elaborated as Algorithm 1 in Heitz et al.~\cite{sggx} .






\section{Inverse Volumetric Rendering}

Now that we have the network structure of NeMF, and regression of the density, abledo, microflake normal, and roughness for each position is inverse volume rendering. We have provided the volumetric rendering functions for NeMF in Sec.~\ref{sec:rwn}, and it is differentiable. We can render from NeMF and use the photometric loss between the rendering result and image observations as supervision for training, i.e.:

\begin{equation}
\mathcal{L}_c = \sum_{\mathbf{r}, I} \| C(\mathbf{r}) - I(\mathbf{r}) \|_2^2
\end{equation}

\noindent where $I(\mathbf{r})$ is the color of ray $\mathbf{r}$ in image $I$. However, only using the photometric loss is not sufficient for producing hight quality results. We add the density field loss $\mathcal{L}_\sigma = \sum_{k}\eta_k (\| \omega_m - \Delta \sigma \|_2^2 +  \max(0,\langle \omega_m, \omega_i \rangle))$ presented in RefNeRF~\cite{nerfactor}, the latent sparsity $\mathcal{L}_z = \sum_{j=1}^{n} \mathrm{KL} (\rho \Vert \hat{\rho_{j}})$ ($\hat{\rho_{j}}$ is the average the $j^{th}$ channel of $\mathbf{z}$ over batch input, $\rho$ is set to 0.05) as in InvRender~\cite{indirect}, for regularization. 
%
%

%
%


We further enforce smoothness on both microflake normals and latent $\mathbf{z}$ using the following smoothness loss:
\begin{equation}
\begin{gathered}  
\mathcal{L}_s = \| \mathcal{N}(\mathcal{F}_g(\mathbf{p})) - \mathcal{N}(\mathcal{F}_g(\mathbf{p}+\epsilon)) \|_1 \\ 
+ \| \mathcal{D}_m(\mathbf{z}) - \mathcal{D}_m(\mathbf{z}+\epsilon) \|_1
\end{gathered}  
\end{equation}
\noindent where $\epsilon$ is a small random variable drawn from a Gaussian distribution with a mean of zero and a variance of 0.01.

The final loss is the sum of all previously defined losses:
\begin{equation}
\mathcal{L}= \mathcal{L}_c + \mathcal{L}_z + \mathcal{L}_{\sigma} + \mathcal{L}_{s}
\end{equation}
\noindent where the corresponding weights of each term are ignored for clear presentation.
\subsection{Illumination and Visibility}

During training, the illumination is unknown and we need to recover simultaneously with the microflake field. We assume that all lights come from an infinitely faraway background and parameterize them as 128-dimensional Spherical Gaussians (SGs), a common practice in inverse rendering.
%
To further improve the quality, we model the visibility of direct illumination for each scene point following InvRender~\cite{indirect} and Relighting4D~\cite{chen2022relighting4d}. 
Specifically, we compute the visibility of each point w.r.t. the light direction $\omega_l$ by marching a ray from the scene to the light in the background and calculating its opacity $\eta$ (Eqn.~\ref{eq:volrender}) with a pre-trained NeRF. We further encode the visibility of environment light w.r.t. a scene point using the SG parameterization, and use an MLP network $\mathcal{V}$ to overfit the visibility SG parameters for each spatial position.
%
%
%
\begin{figure}[t]
	\centering
	\includegraphics[width=1.0 \linewidth]{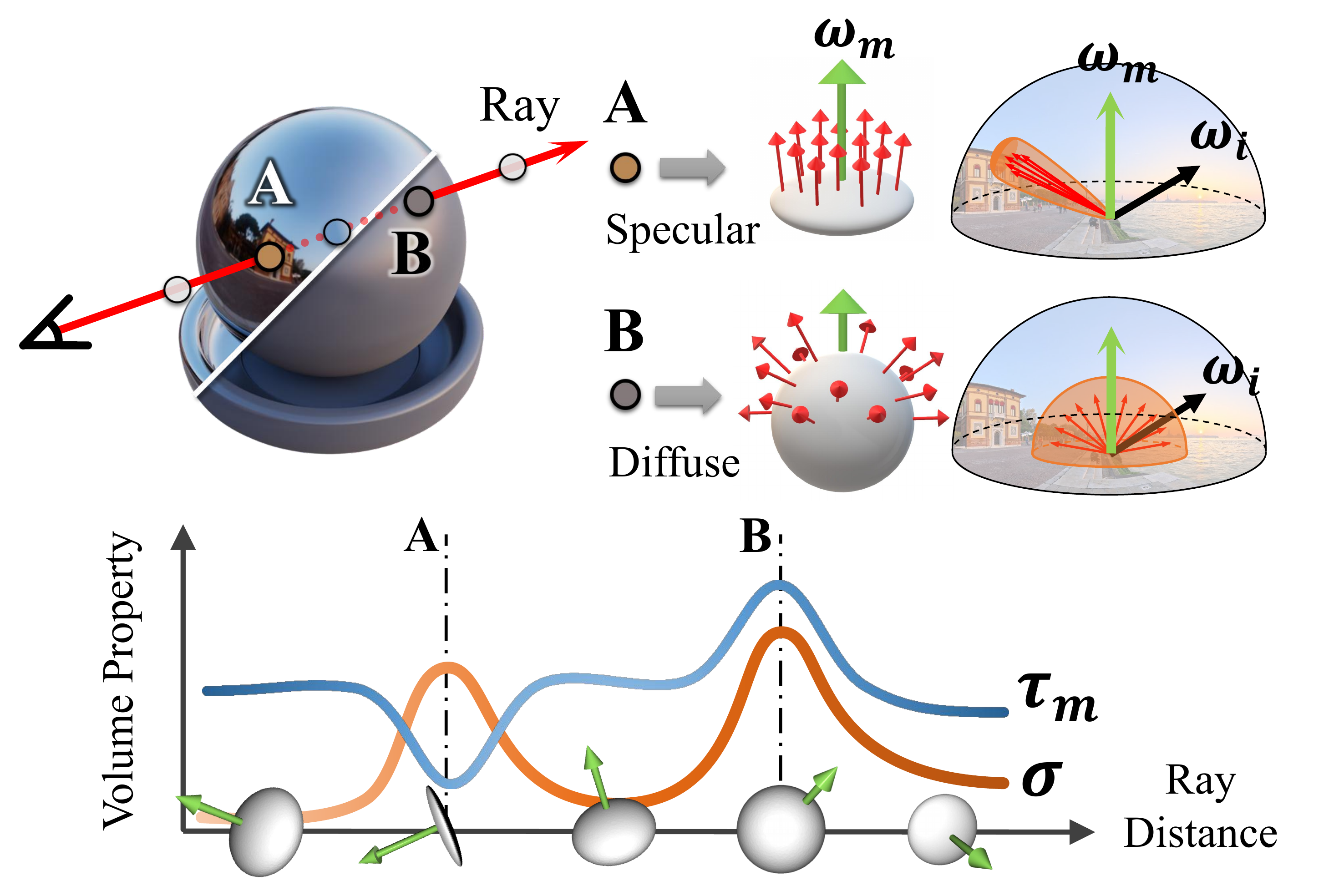}
	
	\caption{
            The rendering process of a ray: point A is relatively specular and point B is relatively diffuse. The ellipsoid that simulates the microflake distribution of A is thiner and that of B is rounder. The final color of ray is integral of transported radiation from the environment light by microflakes along the ray.
	}
	\label{sample}
\end{figure}
%
%
Recall that the radiance at a point $x$, $\nu^*(x, \omega_i)$ contains a diffuse component $\nu(x, \omega_i, f_p^d)$ and a specular component $\nu(x, \omega_i, f_p^s)$. Inspired by previous work~\cite{physg,indirect}, we use the summation of multiplication of environment lighting and visibility as the diffuse component for faster computation:

\begin{equation}
\begin{gathered}
\label{diff-vis}
\nu_d(x) = \frac{a}{\pi} \sum_{\omega_l \in \Omega}  \big \{ [ V(x, \omega_l) Y(\omega_l)] \otimes \\
[L^{\text{sg}}(\omega_l) Y(\omega_l)] \cdot 	\langle \omega_l \cdot \omega_m \rangle \big \}
\end{gathered}
\end{equation}

\noindent where $Y(\omega_l)$ is the SG basis. For the specular reflectance $\nu(x, \omega_i, f_p^s)$, we check the visibility of point $\omega_l$ w.r.t. the light direction during the  ray sampling, and set it to 0 if the light $\omega_l$ is invisible to $x$. We use the visibility adapted color as our final rendered color.

\begin{figure*}[t]
	\centering
	\includegraphics[width=1.0\linewidth]{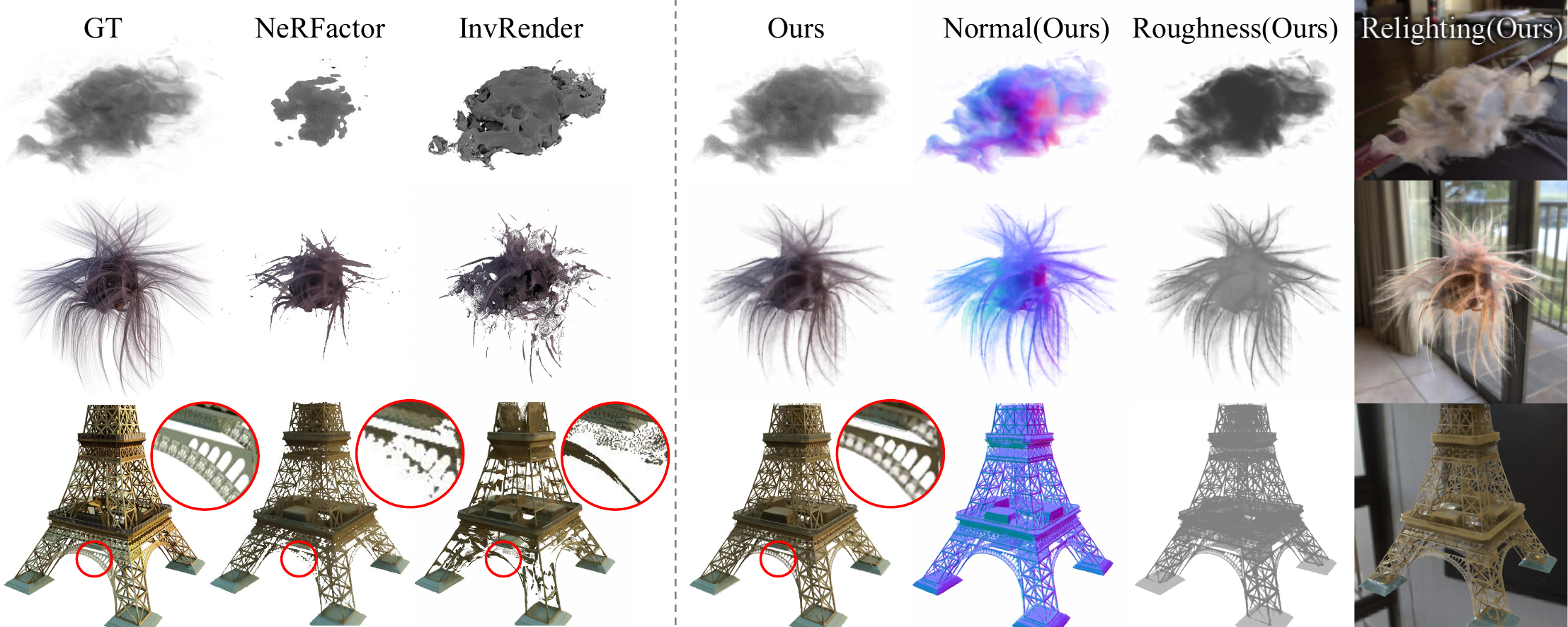}
	\caption{
		\textbf{Comparison on objects with complex geometries} We compare our NeMF with NeRFactor~\cite{nerfactor} and InvRender~\cite{indirect} for synthetic objects with very complicated geometry and material, including a cloud (top), a furry ball (middle), and Eiffel Tower (bottom).
	}
	\label{ablation}
\end{figure*}

\begin{table*}[!htb]
\setlength\tabcolsep{4.5pt}
\centering
\begin{tabular*}{\linewidth}{c c c c c c c c c c c c c c} 
\toprule[1pt]

\multicolumn{3}{c}{\multirow{2}{*}{Method}}& 
Roughness & 
\multicolumn{3}{c}{Albedo} &
\multicolumn{3}{c}{View Synthesis} &
\multicolumn{3}{c}{Relighting}\\

\cmidrule(lr){4-4}\cmidrule(lr){5-7}\cmidrule(lr){8-10}\cmidrule(lr){11-13}

\multicolumn{3}{c}{}&MSE $\downarrow$ 
&PSNR $\uparrow$ &SSIM $\uparrow$ &LPIPS $\downarrow$ 
&PSNR $\uparrow$ &SSIM $\uparrow$ &LPIPS $\downarrow$ 
&PSNR $\uparrow$ &SSIM $\uparrow$ &LPIPS $\downarrow$\\  

\hline
\multicolumn{3}{c}{PhySG\cite{physg}}&-&15.3543&0.8811&0.2463 &15.1026&0.8883&0.2399&14.5316&0.8804&0.2440\\
\multicolumn{3}{c}{NeRFactor\cite{nerfactor}}&-&20.3366&0.9193&0.1093&23.2638&0.9429&0.0997&21.3415&0.9254&0.1056\\
\multicolumn{3}{c}{InvRender\cite{indirect}}&0.0121 & 21.9568 & 0.9339 & \textbf{0.0811} & 27.4874 & 0.9639 & 0.0816 & 24.5309 & 0.9555 & 0.0897\\

\midrule

\multicolumn{3}{c}{Ours}&\textbf{0.0074}&\textbf{23.6982}&\textbf{0.9387}&0.0870&\textbf{28.8920}&\textbf{0.9678}&\textbf{0.0744}&\textbf{25.7048}&\textbf{0.9561}&\textbf{0.0864}\\
\multicolumn{3}{c}{w/o normal}&0.0672&15.4516&0.8914&0.1359&28.1014&0.9627&0.0899&18.6333&0.9134&0.1283\\
\multicolumn{3}{c}{w/o vis.}&0.0445&19.2018&0.9231&0.1057&28.6739&0.9652&0.0834&20.5587&0.9091&0.1339\\
\multicolumn{3}{c}{w/o smoothness}&0.0075&23.5721&0.9360&0.0906&28.8234&0.9638&0.0874&24.9586&0.9445&0.1081\\
\multicolumn{3}{c}{w/o latent space}&0.0160&23.2460&0.9407&0.0930&28.8529&0.9670&0.0775&24.8025&0.9534&0.0945\\
\multicolumn{3}{c}{fewer samples}&0.0527&19.9179&0.9128&0.1099&28.8310&0.9672&0.0771&22.2603&0.9301&0.1170\\

\bottomrule[1pt]
\end{tabular*}
\caption{\textbf{Quantitative evaluations.} We present the average results on the test images of all four synthetic scenes. Though InvRender achieves slightly better LPIPS on albedo estimation, our full model achieves the best performance on roughness estimation, view synthesis and relighting.}
\label{table}
\end{table*}

\vspace{4pt}
With an optimized NeMF, we can relight the captured object by changing the illumination SGs in our microflake volume renderer, edit the material by mapping the albedo to a new color, and create novel volume rendering effects (such as scattering) through changing the weighting factor between diffuse and specular radiance in Eqn.~\ref{eq:radcomb} . 

\begin{figure*}[!htb]
	\centering
	\includegraphics[width=1.0 \linewidth]{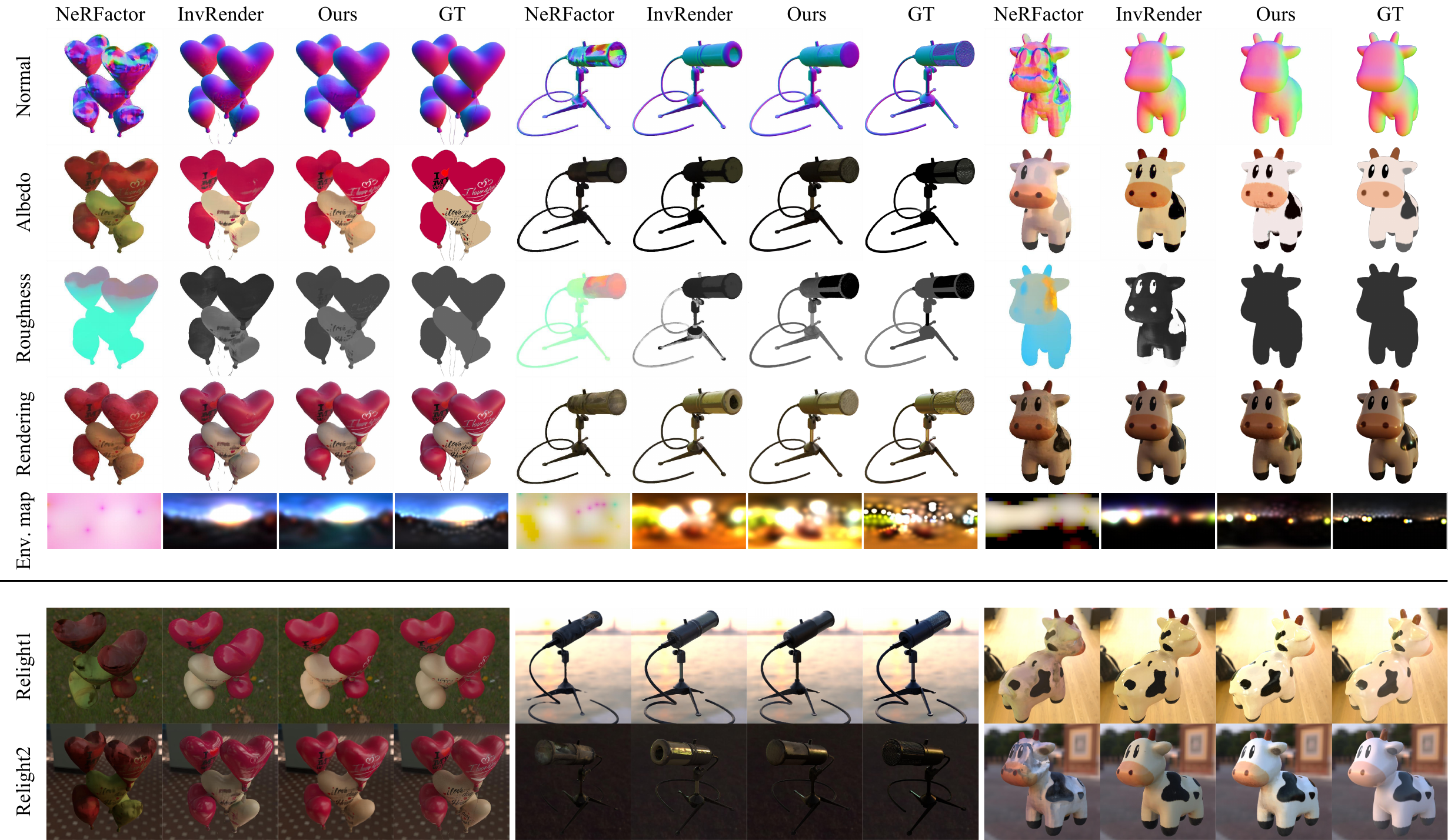}
	
	\caption{
		\textbf{Comparisons with other approaches.} We show the normal, albedo, roughness, and environment map estimated by NeRFactor, InvRender and our method on three scenes. We also compared the rendering results of the new viewpoint under three different lighting conditions (the original light and two novel lights). Note that the roughness of NeRFactor is visualized with the latent code.}
	\label{comparison}
\end{figure*}

\begin{figure}[!htb]
	\centering
	\includegraphics[width=1.0\linewidth]{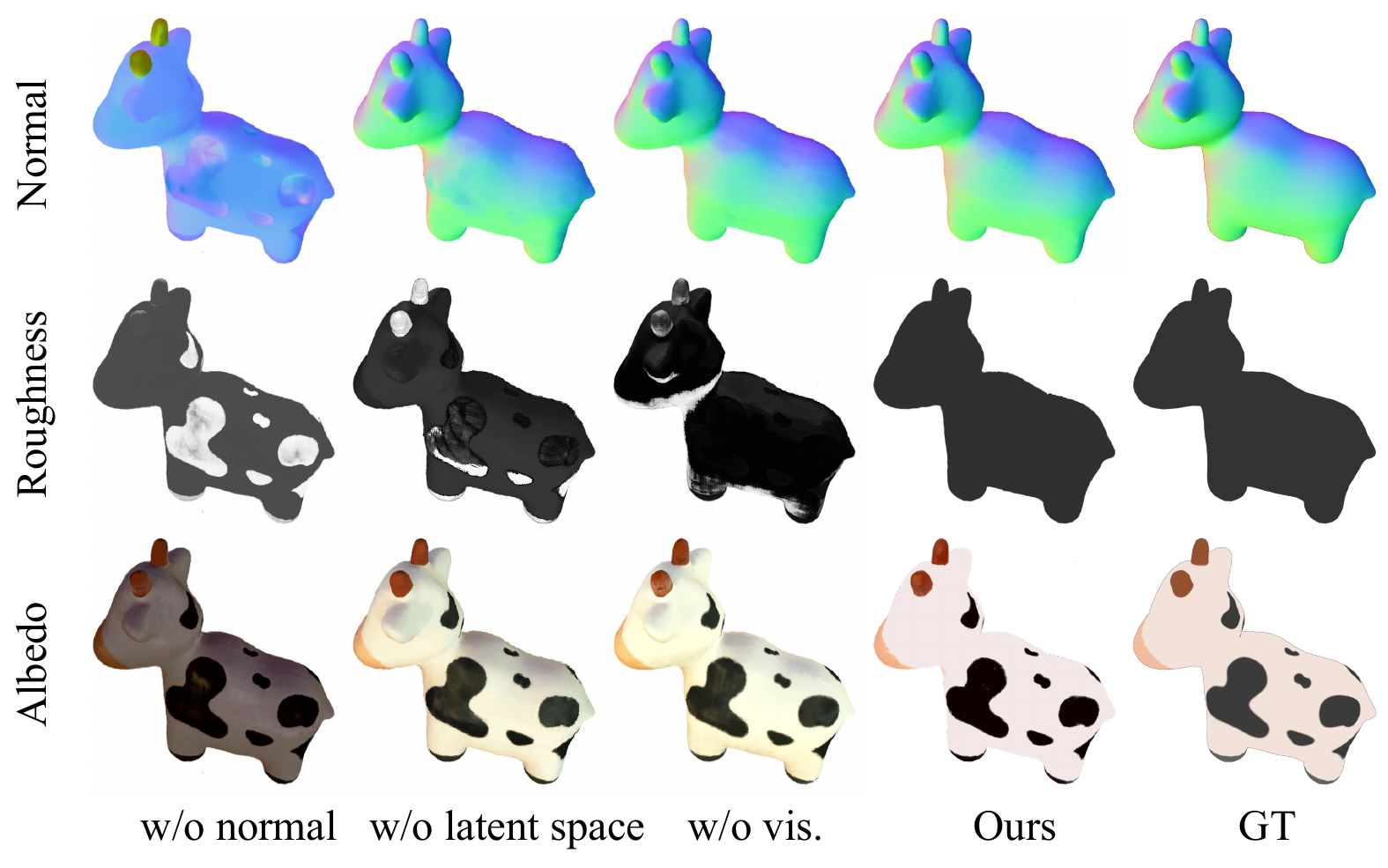}
	
	\caption{
		\textbf{Ablation study on a synthetic scene (Sec. \ref{5.2}}).
	}
	\label{ablation}
\end{figure}

\begin{figure*}[h]
	\centering
	\includegraphics[width=1.0 \linewidth]{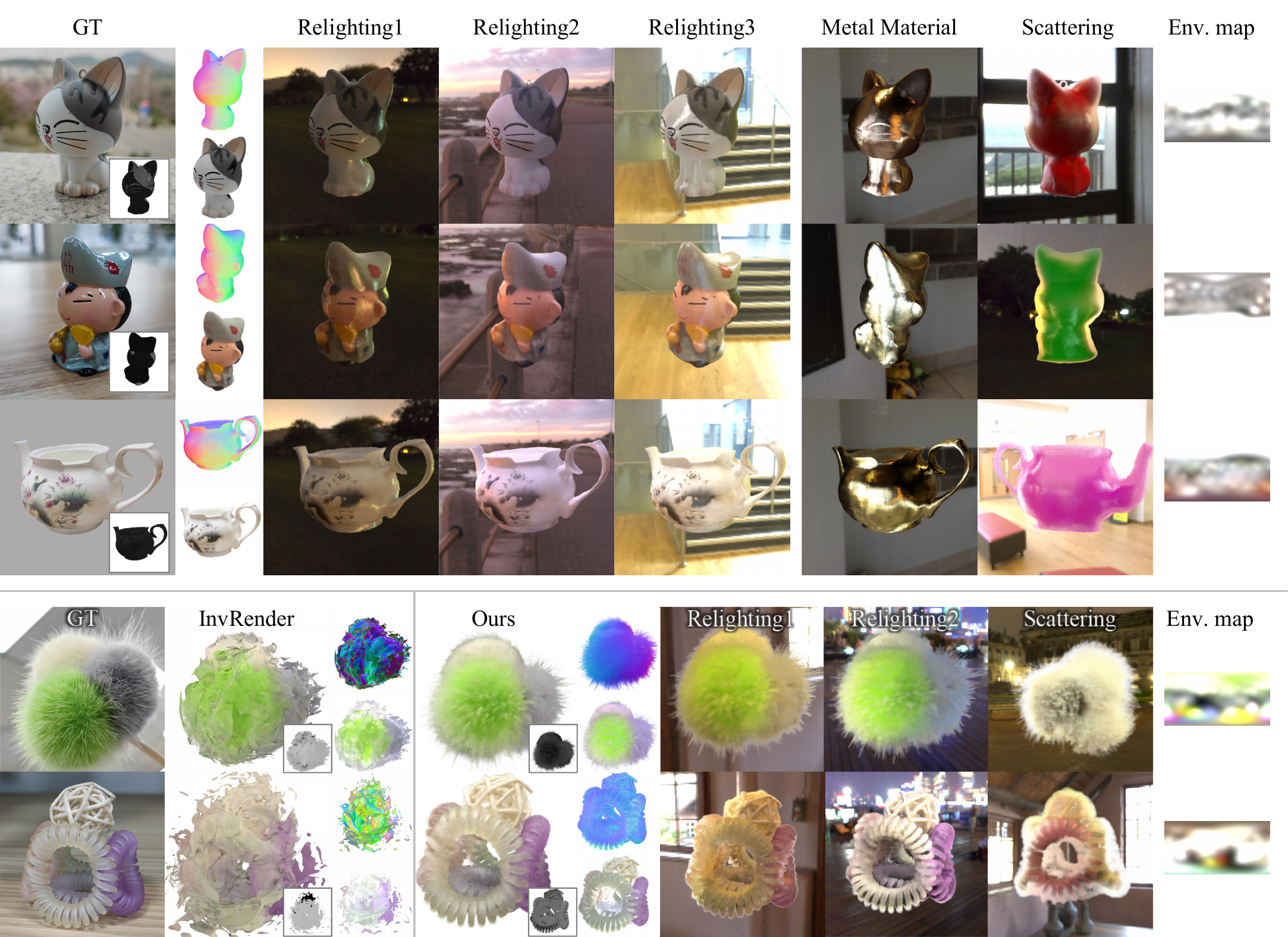}
	
	\caption{\textbf{Results on real captures.}	Our method is capable to handle real-world objects composed of multiple materials. Top three rows, we show results on concrete objects. For each scene, we present the groundtruth, appearance properties, relighting under three novel illuminations, material editing, scattering and environment maps.}
	\label{realdata}
\end{figure*}

\section{Experiments}
We conduct experiments on both synthetic and real-world datasets to evaluate our proposed NeMF. 

\subsection{Synthetic Data}
We use 4 synthetic Blender scenes (balloon, mic, spot, polyhedron) to validate our model. For each object, we choose a specific natural illumination. We render 100 training images for each object under the selected illumination via Blender Cycles, and disable all non-standard post-rendering effects. We also render 200 test images, along with their albedo maps and relighting images using other two environment maps to evaluate the novel view synthesis and relighting performance of our model. All images are in the PNG format and the image resolution is 800x800.

\subsection{Ablation Study}
\label{5.2}
We validate our design choices by ablating 4 major model variants, that are without density and mircoflake normal regularizations, without visibility modeling, without sparsity constraints on the latent code, and without smoothness regularizations. We compare them with our NeMF model to observe whether there is a performance drop quantitatively or qualitatively. We present our quantitative ablation studies in Tab.~\ref{table}, and the qualitative ablation studies in Fig.~\ref{ablation}. 

Fig. ~\ref{ablation} shows that ``w/o normal'' bakes the ambient lighting into the surface of objects, generating the worst results among 4 ablation settings. In ``w/o vis.'', we train a model without calculating the visibility of each location point to constrain our model during pre-training. ``w/o vis.'' assumes the environment illumination is visible to all scene points without considering self-occlusion and interreflection. This setup leads to incorrect predictions, especially around the occlusion boundaries. ``w/o latent space'' maps location points to parameters of neural network straightforwardly, instead of transforming it to a latent space first. This results in incorrect extraction of the object’s materials from images, as well as vaguer edges compared to our model.

\subsection{Comparison}
We compare our NeMF with 3 baselines (NeRFactor~\cite{nerfactor}, PhySG~\cite{physg} and InvRender~\cite{indirect}) in the tasks of novel view synthesis and relighting on the synthetic datasets. We use Peak Signal-to-Noise Ratio (PSNR) and Structural Similarity Index Measure (SSIM) as metrics. We also use Learned Perceptual Image Patch Similarity (LPIPS)\cite{lpips}, where lower is better. Tab.~\ref{table} demonstrates that there is a bigger gap between PhySG's results and ground truth. The main reason is that PhySG assumes the object recovered is homogeneous, therefore it cannot even recover the shapes of objects with multiple materials, the mic for example.Tab.~\ref{table} and Fig.~\ref{comparison} show that our model has a better performance both quantitatively and qualitatively than NeRFactor. Fig.~\ref{comparison} demonstrates that NeRFactor cannot extract an accurate environment map, and has a lower performance on the balloon datasets especially, as the albedo is baked into the environment map and thus leads to poor relighting results. Compared to NeRFactor, InvRender and our model have much smoother results on normal estimation, which are closer to ground truth.As shown in Fig.~\ref{comparison} and Tab.~\ref{table}, although InvRender has more access to ground truth on both novel view synthesis and relighting than previous work, it fails to recover some details of scenes compared to our model. The results also demonstrate that InvRender may not be able to decompose the scene into geometry and illumination accurately when the surface of a scene is complex.

\subsection{Real Data Experiments}
To demonstrate that our approach is able to handle real-work data, we test on 5 datasets including the `cat', `monk', `teapot' as concrete and simple objects, and two complicate scenes, `furry ball' and `bands'. We capture all `cat', `monk', `furry ball' and `bands' scenes with a mobile phone, OPPO RENO5, and record a video while the collector walking around the object. The frames and the resolution of each video are 30fps and 960x540, respectively. We extract around 400 frames from the video for each object, and from which we randomly select 100 images for training. We obtain the camera parameters using COLMAP\cite{colmap} (model: "PINHOLE"). Before training, we remove the background of all images using background removal tools~\cite{removebg}. We illustrate the relighting, material editing and volume scattering results in Fig.~\ref{realdata}. As we can see, the relighting result is faithful, given that we capture the data with a mobile phone in office environment lighting. The metal-like effect of the `cat' and `monk' is very realistic. 
While generating volume scattering effect, we change the object color while preserving the specular terms and give the object a semi-translucent appearance, which is infeasible for most surface based inverse rendering approaches. Particularly, our NeMF can handle the complicate objects very well, as shown bottom two rows in Fig.~\ref{realdata}.

\section{Conclusion}

In this paper, we present the Neural Microflake Field (NeMF), which uses an implicit coordinate network to encode a microflake volume for inverse volumetric rendering. Our NeMF is a fully volumetric model and can well handle complicate geometries and materials. 
Our NeMF still has several limitations: the recovered geometry is not as smooth as the neural surface-based methods, we adopt
a per-scene optimization scheme for training which is not generalizable, our approach requires a large number of input images (over 100), and both training and rendering are time-intensive. In the future, we plan to address above issues by incorporating recent advances, including the multi-resolution hashing technique in InstantNGP~\cite{muller2022instant}, and factorization methods~\cite{chen2022tensorf, chan2022efficient}. We would like to further adapt our trained network to parameterizations (PlenOctree~\cite{yu2021plenoctrees}) for real-time rendering and relighting.

{\small
\bibliographystyle{ieee_fullname}
\bibliography{main}

\begin{thebibliography}{10}\itemsep=-1pt

\bibitem{removebg}
1.
\newblock Natural population growth rate.
\newblock \url{https://www.remove.bg/}.
\newblock Accessed: October 1, 2022.

\bibitem{deepsvbrdf}
Louis-Philippe Asselin, Denis Laurendeau, and Jean-Francois Lalonde.
\newblock Deep svbrdf estimation on real materials.
\newblock In {\em 2020 International Conference on 3D Vision (3DV)}, pages
  1157--1166. IEEE, 2020.

\bibitem{barron2014shape}
Jonathan~T Barron and Jitendra Malik.
\newblock Shape, illumination, and reflectance from shading.
\newblock {\em IEEE transactions on pattern analysis and machine intelligence},
  37(8):1670--1687, 2014.

\bibitem{bi2020neural}
Sai Bi, Zexiang Xu, Pratul Srinivasan, Ben Mildenhall, Kalyan Sunkavalli,
  Milo{\v{s}} Ha{\v{s}}an, Yannick Hold-Geoffroy, David Kriegman, and Ravi
  Ramamoorthi.
\newblock Neural reflectance fields for appearance acquisition.
\newblock {\em arXiv preprint arXiv:2008.03824}, 2020.

\bibitem{bi2020deepref}
Sai Bi, Zexiang Xu, Kalyan Sunkavalli, Milo{\v{s}} Ha{\v{s}}an, Yannick
  Hold-Geoffroy, David Kriegman, and Ravi Ramamoorthi.
\newblock Deep reflectance volumes: Relightable reconstructions from multi-view
  photometric images.
\newblock In {\em European Conference on Computer Vision}, pages 294--311.
  Springer, 2020.

\bibitem{bi2020deep}
Sai Bi, Zexiang Xu, Kalyan Sunkavalli, David Kriegman, and Ravi Ramamoorthi.
\newblock Deep 3d capture: Geometry and reflectance from sparse multi-view
  images.
\newblock In {\em Proceedings of the IEEE/CVF Conference on Computer Vision and
  Pattern Recognition}, pages 5960--5969, 2020.

\bibitem{nerd}
Mark Boss, Raphael Braun, Varun Jampani, Jonathan~T Barron, Ce Liu, and Hendrik
  Lensch.
\newblock Nerd: Neural reflectance decomposition from image collections.
\newblock In {\em Proceedings of the IEEE/CVF International Conference on
  Computer Vision}, pages 12684--12694, 2021.

\bibitem{deepbrdf}
Mark Boss, Fabian Groh, Sebastian Herholz, and Hendrik~PA Lensch.
\newblock Deep dual loss brdf parameter estimation.
\newblock In {\em MAM@ EGSR}, pages 41--44, 2018.

\bibitem{chan2022efficient}
Eric~R Chan, Connor~Z Lin, Matthew~A Chan, Koki Nagano, Boxiao Pan, Shalini
  De~Mello, Orazio Gallo, Leonidas~J Guibas, Jonathan Tremblay, Sameh Khamis,
  et~al.
\newblock Efficient geometry-aware 3d generative adversarial networks.
\newblock In {\em Proceedings of the IEEE/CVF Conference on Computer Vision and
  Pattern Recognition}, pages 16123--16133, 2022.

\bibitem{chen2022tensorf}
Anpei Chen, Zexiang Xu, Andreas Geiger, Jingyi Yu, and Hao Su.
\newblock Tensorf: Tensorial radiance fields.
\newblock In {\em Computer Vision--ECCV 2022: 17th European Conference, Tel
  Aviv, Israel, October 23--27, 2022, Proceedings, Part XXXII}, pages 333--350.
  Springer, 2022.

\bibitem{chen2020neural}
Zhang Chen, Anpei Chen, Guli Zhang, Chengyuan Wang, Yu Ji, Kiriakos~N
  Kutulakos, and Jingyi Yu.
\newblock A neural rendering framework for free-viewpoint relighting.
\newblock In {\em Proceedings of the IEEE/CVF Conference on Computer Vision and
  Pattern Recognition}, pages 5599--5610, 2020.

\bibitem{chen2022relighting4d}
Zhaoxi Chen and Ziwei Liu.
\newblock Relighting4d: Neural relightable human from videos.
\newblock In {\em European Conference on Computer Vision}, pages 606--623.
  Springer, 2022.

\bibitem{debevec2000acquiring}
Paul Debevec, Tim Hawkins, Chris Tchou, Haarm-Pieter Duiker, Westley Sarokin,
  and Mark Sagar.
\newblock Acquiring the reflectance field of a human face.
\newblock In {\em Proceedings of the 27th annual conference on Computer
  graphics and interactive techniques}, pages 145--156, 2000.

\bibitem{dong2014appearance}
Yue Dong, Guojun Chen, Pieter Peers, Jiawan Zhang, and Xin Tong.
\newblock Appearance-from-motion: Recovering spatially varying surface
  reflectance under unknown lighting.
\newblock {\em ACM Transactions on Graphics (TOG)}, 33(6):1--12, 2014.

\bibitem{drebin1988volume}
Robert~A Drebin, Loren Carpenter, and Pat Hanrahan.
\newblock Volume rendering.
\newblock {\em ACM Siggraph Computer Graphics}, 22(4):65--74, 1988.

\bibitem{gao2020deferred}
Duan Gao, Guojun Chen, Yue Dong, Pieter Peers, Kun Xu, and Xin Tong.
\newblock Deferred neural lighting: free-viewpoint relighting from unstructured
  photographs.
\newblock {\em ACM Transactions on Graphics (TOG)}, 39(6):1--15, 2020.

\bibitem{georgoulis2015gaussian}
Stamatios Georgoulis, Vincent Vanweddingen, Marc Proesmans, and Luc Van~Gool.
\newblock A gaussian process latent variable model for brdf inference.
\newblock In {\em Proceedings of the IEEE International Conference on Computer
  Vision}, pages 3559--3567, 2015.

\bibitem{gkioulekas2013inverse}
Ioannis Gkioulekas, Shuang Zhao, Kavita Bala, Todd Zickler, and Anat Levin.
\newblock Inverse volume rendering with material dictionaries.
\newblock {\em ACM Transactions on Graphics (TOG)}, 32(6):1--13, 2013.

\bibitem{godard2015multi}
Clement Godard, Peter Hedman, Wenbin Li, and Gabriel~J Brostow.
\newblock Multi-view reconstruction of highly specular surfaces in uncontrolled
  environments.
\newblock In {\em 2015 International Conference on 3D Vision}, pages 19--27.
  IEEE, 2015.

\bibitem{goel2020shape}
Purvi Goel, Loudon Cohen, James Guesman, Vikas Thamizharasan, James Tompkin,
  and Daniel Ritchie.
\newblock Shape from tracing: Towards reconstructing 3d object geometry and
  svbrdf material from images via differentiable path tracing.
\newblock In {\em 2020 International Conference on 3D Vision (3DV)}, pages
  1186--1195. IEEE, 2020.

\bibitem{guo2019relightables}
Kaiwen Guo, Peter Lincoln, Philip Davidson, Jay Busch, Xueming Yu, Matt Whalen,
  Geoff Harvey, Sergio Orts-Escolano, Rohit Pandey, Jason Dourgarian, et~al.
\newblock The relightables: Volumetric performance capture of humans with
  realistic relighting.
\newblock {\em ACM Transactions on Graphics (ToG)}, 38(6):1--19, 2019.

\bibitem{sggx}
Eric Heitz, Jonathan Dupuy, Cyril Crassin, and Carsten Dachsbacher.
\newblock The sggx microflake distribution.
\newblock {\em ACM Transactions on Graphics (TOG)}, 34(4):1--11, 2015.

\bibitem{jakob2010radiative}
Wenzel Jakob, Adam Arbree, Jonathan~T Moon, Kavita Bala, and Steve Marschner.
\newblock A radiative transfer framework for rendering materials with
  anisotropic structure.
\newblock In {\em ACM SIGGRAPH 2010 papers}, pages 1--13. 2010.

\bibitem{jiang2020sdfdiff}
Yue Jiang, Dantong Ji, Zhizhong Han, and Matthias Zwicker.
\newblock Sdfdiff: Differentiable rendering of signed distance fields for 3d
  shape optimization.
\newblock In {\em Proceedings of the IEEE/CVF conference on computer vision and
  pattern recognition}, pages 1251--1261, 2020.

\bibitem{efficientbrdf}
Jason Lawrence, Szymon Rusinkiewicz, and Ravi Ramamoorthi.
\newblock Efficient brdf importance sampling using a factored representation.
\newblock {\em ACM Transactions on Graphics (ToG)}, 23(3):496--505, 2004.

\bibitem{image-based}
Hendrik Lensch, Jan Kautz, Michael Goesele, Wolfgang Heidrich, and Hans-Peter
  Seidel.
\newblock Image-based reconstruction of spatially varying materials.
\newblock In {\em Eurographics Workshop on Rendering Techniques}, pages
  103--114. Springer, 2001.

\bibitem{lensch2003image}
Hendrik~PA Lensch, Jan Kautz, Michael Goesele, Wolfgang Heidrich, and
  Hans-Peter Seidel.
\newblock Image-based reconstruction of spatial appearance and geometric
  detail.
\newblock {\em ACM Transactions on Graphics (TOG)}, 22(2):234--257, 2003.

\bibitem{planned}
Hendrik~PA Lensch, Jochen Lang, Asla~M S{\'a}, and Hans-Peter Seidel.
\newblock Planned sampling of spatially varying brdfs.
\newblock In {\em Computer graphics forum}, volume~22, pages 473--482. Wiley
  Online Library, 2003.

\bibitem{li2020inverse}
Zhengqin Li, Mohammad Shafiei, Ravi Ramamoorthi, Kalyan Sunkavalli, and
  Manmohan Chandraker.
\newblock Inverse rendering for complex indoor scenes: Shape, spatially-varying
  lighting and svbrdf from a single image.
\newblock In {\em Proceedings of the IEEE/CVF Conference on Computer Vision and
  Pattern Recognition}, pages 2475--2484, 2020.

\bibitem{li2018learning}
Zhengqin Li, Zexiang Xu, Ravi Ramamoorthi, Kalyan Sunkavalli, and Manmohan
  Chandraker.
\newblock Learning to reconstruct shape and spatially-varying reflectance from
  a single image.
\newblock {\em ACM Transactions on Graphics (TOG)}, 37(6):1--11, 2018.

\bibitem{lichy2021shape}
Daniel Lichy, Jiaye Wu, Soumyadip Sengupta, and David~W Jacobs.
\newblock Shape and material capture at home.
\newblock In {\em Proceedings of the IEEE/CVF Conference on Computer Vision and
  Pattern Recognition}, pages 6123--6133, 2021.

\bibitem{maier2017intrinsic3d}
Robert Maier, Kihwan Kim, Daniel Cremers, Jan Kautz, and Matthias Nie{\ss}ner.
\newblock Intrinsic3d: High-quality 3d reconstruction by joint appearance and
  geometry optimization with spatially-varying lighting.
\newblock In {\em Proceedings of the IEEE international conference on computer
  vision}, pages 3114--3122, 2017.

\bibitem{mescheder2019occupancy}
Lars Mescheder, Michael Oechsle, Michael Niemeyer, Sebastian Nowozin, and
  Andreas Geiger.
\newblock Occupancy networks: Learning 3d reconstruction in function space.
\newblock In {\em Proceedings of the IEEE/CVF conference on computer vision and
  pattern recognition}, pages 4460--4470, 2019.

\bibitem{nerf}
Ben Mildenhall, Pratul~P Srinivasan, Matthew Tancik, Jonathan~T Barron, Ravi
  Ramamoorthi, and Ren Ng.
\newblock Nerf: Representing scenes as neural radiance fields for view
  synthesis.
\newblock {\em Communications of the ACM}, 65(1):99--106, 2021.

\bibitem{muller2022instant}
Thomas M{\"u}ller, Alex Evans, Christoph Schied, and Alexander Keller.
\newblock Instant neural graphics primitives with a multiresolution hash
  encoding.
\newblock {\em ACM Transactions on Graphics (ToG)}, 41(4):1--15, 2022.

\bibitem{nam2018practical}
Giljoo Nam, Joo~Ho Lee, Diego Gutierrez, and Min~H Kim.
\newblock Practical svbrdf acquisition of 3d objects with unstructured flash
  photography.
\newblock {\em ACM Transactions on Graphics (TOG)}, 37(6):1--12, 2018.

\bibitem{niemeyer2020differentiable}
Michael Niemeyer, Lars Mescheder, Michael Oechsle, and Andreas Geiger.
\newblock Differentiable volumetric rendering: Learning implicit 3d
  representations without 3d supervision.
\newblock In {\em Proceedings of the IEEE/CVF Conference on Computer Vision and
  Pattern Recognition}, pages 3504--3515, 2020.

\bibitem{nimier2022unbiased}
Merlin Nimier-David, Thomas M{\"u}ller, Alexander Keller, and Wenzel Jakob.
\newblock Unbiased inverse volume rendering with differential trackers.
\newblock {\em ACM Transactions on Graphics (TOG)}, 41(4):1--20, 2022.

\bibitem{oxholm2014multiview}
Geoffrey Oxholm and Ko Nishino.
\newblock Multiview shape and reflectance from natural illumination.
\newblock In {\em Proceedings of the IEEE Conference on Computer Vision and
  Pattern Recognition}, pages 2155--2162, 2014.

\bibitem{park2019deepsdf}
Jeong~Joon Park, Peter Florence, Julian Straub, Richard Newcombe, and Steven
  Lovegrove.
\newblock Deepsdf: Learning continuous signed distance functions for shape
  representation.
\newblock In {\em Proceedings of the IEEE/CVF conference on computer vision and
  pattern recognition}, pages 165--174, 2019.

\bibitem{park2020seeing}
Jeong~Joon Park, Aleksander Holynski, and Steven~M Seitz.
\newblock Seeing the world in a bag of chips.
\newblock In {\em Proceedings of the IEEE/CVF Conference on Computer Vision and
  Pattern Recognition}, pages 1417--1427, 2020.

\bibitem{philip2019multi}
Julien Philip, Micha{\"e}l Gharbi, Tinghui Zhou, Alexei~A Efros, and George
  Drettakis.
\newblock Multi-view relighting using a geometry-aware network.
\newblock {\em ACM Trans. Graph.}, 38(4):78--1, 2019.

\bibitem{sang2020single}
Shen Sang and Manmohan Chandraker.
\newblock Single-shot neural relighting and svbrdf estimation.
\newblock In {\em European Conference on Computer Vision}, pages 85--101.
  Springer, 2020.

\bibitem{schmitt2020joint}
Carolin Schmitt, Simon Donne, Gernot Riegler, Vladlen Koltun, and Andreas
  Geiger.
\newblock On joint estimation of pose, geometry and svbrdf from a handheld
  scanner.
\newblock In {\em Proceedings of the IEEE/CVF Conference on Computer Vision and
  Pattern Recognition}, pages 3493--3503, 2020.

\bibitem{colmap}
Johannes~L Schonberger and Jan-Michael Frahm.
\newblock Structure-from-motion revisited.
\newblock In {\em Proceedings of the IEEE conference on computer vision and
  pattern recognition}, pages 4104--4113, 2016.

\bibitem{sengupta2019neural}
Soumyadip Sengupta, Jinwei Gu, Kihwan Kim, Guilin Liu, David~W Jacobs, and Jan
  Kautz.
\newblock Neural inverse rendering of an indoor scene from a single image.
\newblock In {\em Proceedings of the IEEE/CVF International Conference on
  Computer Vision}, pages 8598--8607, 2019.

\bibitem{sitzmann2020metasdf}
Vincent Sitzmann, Eric Chan, Richard Tucker, Noah Snavely, and Gordon
  Wetzstein.
\newblock Metasdf: Meta-learning signed distance functions.
\newblock {\em Advances in Neural Information Processing Systems},
  33:10136--10147, 2020.

\bibitem{nerv}
Pratul~P Srinivasan, Boyang Deng, Xiuming Zhang, Matthew Tancik, Ben
  Mildenhall, and Jonathan~T Barron.
\newblock Nerv: Neural reflectance and visibility fields for relighting and
  view synthesis.
\newblock In {\em Proceedings of the IEEE/CVF Conference on Computer Vision and
  Pattern Recognition}, pages 7495--7504, 2021.

\bibitem{vicini2021path}
Delio Vicini, S{\'e}bastien Speierer, and Wenzel Jakob.
\newblock Path replay backpropagation: differentiating light paths using
  constant memory and linear time.
\newblock {\em ACM Transactions on Graphics (TOG)}, 40(4):1--14, 2021.

\bibitem{wei2020object}
Xin Wei, Guojun Chen, Yue Dong, Stephen Lin, and Xin Tong.
\newblock Object-based illumination estimation with rendering-aware neural
  networks.
\newblock In {\em European Conference on Computer Vision}, pages 380--396.
  Springer, 2020.

\bibitem{xia2016recovering}
Rui Xia, Yue Dong, Pieter Peers, and Xin Tong.
\newblock Recovering shape and spatially-varying surface reflectance under
  unknown illumination.
\newblock {\em ACM Transactions on Graphics (TOG)}, 35(6):1--12, 2016.

\bibitem{yariv2020multiview}
Lior Yariv, Yoni Kasten, Dror Moran, Meirav Galun, Matan Atzmon, Basri Ronen,
  and Yaron Lipman.
\newblock Multiview neural surface reconstruction by disentangling geometry and
  appearance.
\newblock {\em Advances in Neural Information Processing Systems},
  33:2492--2502, 2020.

\bibitem{yu2021plenoctrees}
Alex Yu, Ruilong Li, Matthew Tancik, Hao Li, Ren Ng, and Angjoo Kanazawa.
\newblock Plenoctrees for real-time rendering of neural radiance fields.
\newblock In {\em Proceedings of the IEEE/CVF International Conference on
  Computer Vision}, pages 5752--5761, 2021.

\bibitem{yu2019inverserendernet}
Ye Yu and William~AP Smith.
\newblock Inverserendernet: Learning single image inverse rendering.
\newblock In {\em Proceedings of the IEEE/CVF Conference on Computer Vision and
  Pattern Recognition}, pages 3155--3164, 2019.

\bibitem{zhang2021path}
Cheng Zhang, Zihan Yu, and Shuang Zhao.
\newblock Path-space differentiable rendering of participating media.
\newblock {\em ACM Transactions on Graphics (TOG)}, 40(4):1--15, 2021.

\bibitem{IRON}
Kai Zhang, Fujun Luan, Zhengqi Li, and Noah Snavely.
\newblock Iron: Inverse rendering by optimizing neural sdfs and materials from
  photometric images.
\newblock In {\em Proceedings of the IEEE/CVF Conference on Computer Vision and
  Pattern Recognition}, pages 5565--5574, 2022.

\bibitem{physg}
Kai Zhang, Fujun Luan, Qianqian Wang, Kavita Bala, and Noah Snavely.
\newblock Physg: Inverse rendering with spherical gaussians for physics-based
  material editing and relighting.
\newblock In {\em Proceedings of the IEEE/CVF Conference on Computer Vision and
  Pattern Recognition}, pages 5453--5462, 2021.

\bibitem{lpips}
Richard Zhang, Phillip Isola, Alexei~A Efros, Eli Shechtman, and Oliver Wang.
\newblock The unreasonable effectiveness of deep features as a perceptual
  metric.
\newblock In {\em Proceedings of the IEEE conference on computer vision and
  pattern recognition}, pages 586--595, 2018.

\bibitem{zhang2021neural}
Xiuming Zhang, Sean Fanello, Yun-Ta Tsai, Tiancheng Sun, Tianfan Xue, Rohit
  Pandey, Sergio Orts-Escolano, Philip Davidson, Christoph Rhemann, Paul
  Debevec, et~al.
\newblock Neural light transport for relighting and view synthesis.
\newblock {\em ACM Transactions on Graphics (TOG)}, 40(1):1--17, 2021.

\bibitem{nerfactor}
Xiuming Zhang, Pratul~P Srinivasan, Boyang Deng, Paul Debevec, William~T
  Freeman, and Jonathan~T Barron.
\newblock Nerfactor: Neural factorization of shape and reflectance under an
  unknown illumination.
\newblock {\em ACM Transactions on Graphics (TOG)}, 40(6):1--18, 2021.

\bibitem{indirect}
Yuanqing Zhang, Jiaming Sun, Xingyi He, Huan Fu, Rongfei Jia, and Xiaowei Zhou.
\newblock Modeling indirect illumination for inverse rendering.
\newblock In {\em Proceedings of the IEEE/CVF Conference on Computer Vision and
  Pattern Recognition}, pages 18643--18652, 2022.

\bibitem{zollhofer2015shading}
Michael Zollh{\"o}fer, Angela Dai, Matthias Innmann, Chenglei Wu, Marc
  Stamminger, Christian Theobalt, and Matthias Nie{\ss}ner.
\newblock Shading-based refinement on volumetric signed distance functions.
\newblock {\em ACM Transactions on Graphics (TOG)}, 34(4):1--14, 2015.

\end{thebibliography}
}

\end{document}